\documentclass[11pt,a4paper]{article}
\usepackage[hyperref]{eacl2021}
\usepackage{times}
\usepackage{latexsym}
\usepackage{url}
\usepackage{amsmath}
\usepackage{multirow}
\usepackage{bbm}
\usepackage{graphicx}
\usepackage{amssymb}
\usepackage{booktabs}
\usepackage{adjustbox}
\usepackage{xspace}
\usepackage{algpseudocode}
\usepackage{algorithm}

\newcommand{\rstrel}[1]{\texttt{#1}\xspace}

\newcommand{\secref}[2][]{Section#1~\ref{sec:#2}}

\newcommand{\tabref}[2][]{Table#1~\ref{tab:#2}}
\newcommand{\figref}[2][]{Figure#1~\ref{fig:#2}}

\newcommand{\eqnref}[2][]{Equation#1~(\ref{eqn:#2})}
\newcommand{\appref}[2][]{Appendix#1~\ref{#2}}
\newcommand{\algoref}[2][]{Algorithm#1~\ref{alg:#2}}

\newcommand{\sentfeat}{\ensuremath{\dagger}}
\newcommand{\parafeat}{\ensuremath{\ddagger}}

\usepackage{microtype}

\aclfinalcopy % Uncomment this line for the final submission
\title{Top-down Discourse Parsing via Sequence Labelling}

\author{Fajri Koto \qquad Jey Han Lau \qquad Timothy Baldwin\\
	School of Computing and Information Systems \\
	The University of Melbourne \\
	\texttt{\small ffajri@student.unimelb.edu.au, jeyhan.lau@gmail.com,
		tbaldwin@unimelb.edu.au} \\
}

\date{}

\begin{document}
\maketitle
\begin{abstract}

%Traditionally, studies in RST discourse parsing are dominated by the
%bottom-up paradigm, implemented using transition-based methods or the
%CYK algorithm. To construct the discourse tree, transition-based parsers,
%for instance, decompose the task into a series of actions such as
%\texttt{SHIFT} and \texttt{REDUCE} using a stack/queue.

We introduce a top-down approach to discourse parsing that is
conceptually simpler than its predecessors \cite{kobayashi2020top,zhang-etal-2020-top}. By framing the task as
a sequence labelling problem where the goal is to iteratively segment a
document into individual discourse units, we are able to eliminate the decoder and reduce the search space for splitting points. We explore both traditional recurrent models and modern pre-trained transformer models for the task, and additionally introduce a novel dynamic oracle for top-down parsing.
Based on the \texttt{Full} metric, our proposed LSTM model sets a
new state-of-the-art for RST parsing.\footnote{Code and trained models: \url{https://github.com/fajri91/NeuralRST-TopDown}}

%  We introduce a top-down approach to RST parsing. introduce a novel
%  method to construct a neural RST
%  discourse tree in a top-down manner. Although discourse parsing has
%  primarily been studied via bottom-up approaches such as
%  transition-based parsers or the CYK algorithm, we argue that these models are
%  over-complicated.  We hypothesize that these such complexities limit
%  the ability of the model to predict the correct RST tree, and that
%  straightforward segmentation via an RNN or transformer
%  simplifies this process and will lead to better results. We introduce our top-down algorithm as a
%  divide-and-conquer scheme where all EDUs are segmented sequentially
%  until the leaves are reached. By introducing an element penalty and new
%  dynamic oracle for segmentation, we empirically show that the
%  top-down approach using an RNN substantially outperforms all baselines.

\end{abstract}

\section{Introduction}

Discourse analysis involves the modelling of the structure of text in a
document. It provides a systematic way to understand how texts are
segmented hierarchically into discourse units, and the relationships
between them. Unlike syntax parsing which models the relationship of
words in a sentence, discourse parsing operates at the document-level,
and aims to explain the flow of writing. Studies have found that
discourse parsing is beneficial for downstream NLP tasks including
document-level sentiment analysis \cite{bhatia2015better} and
abstractive summarization \cite{koto2019improved}.

Rhetorical Structure Theory (RST; \citet{mann1988rhet}) is one of the
most widely used discourse theories in NLP
\cite{hernault2010hilda, feng-hirst-2014-linear,
ji-eisenstein-2014-representation, li-etal-2016-discourse,
yu2018transition}. RST organizes text spans into a tree, where the
leaves represent the basic unit of discourse, known as elementary
discourse units (EDUs). EDUs are typically clauses of a sentence.
Non-terminal nodes in the tree represent discourse unit relations.

\begin{figure}
	\centering
	\includegraphics[width=2.5in]{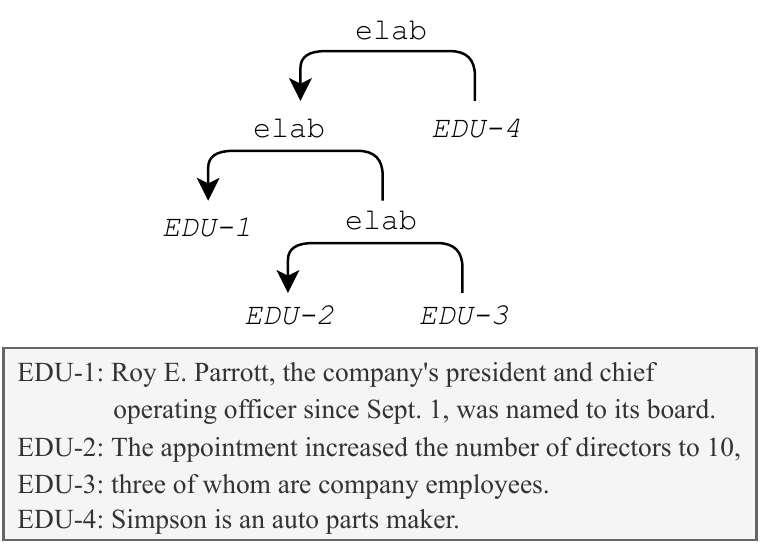}
	\caption{An example discourse tree, from the RST Discourse Treebank (\rstrel{elab} = elaboration).}
	\label{fig:rsttree}
\end{figure}

In \figref{rsttree}, we present an example RST tree with
four EDUs spanning two sentences. In this discourse tree, EDUs are
hierarchically connected with arrows and the discourse label \texttt{elab}.
The direction of arrows indicates the nuclearity of relations, wherein a
``satellite'' points to its ``nucleus''.  The {satellite} unit is a
supporting sentence for the {nucleus} unit and contains less prominent
information. It is standard practice that the RST tree is trained and
evaluated in a right-heavy binarized manner, resulting in three forms of
binary nuclearity relationships between EDUs: {Nucleus--Satellite}, {Satellite--Nucleus},
and  {Nucleus--Nucleus}. In this work, eighteen coarse-grained relations are
considered as discourse labels, consistent with earlier work \cite{yu-etal-2018-transition}.\footnote{Details of individual relations can be found at: \url{http://www.sfu.ca/rst/index.html}}
%: \rstrel{purp},
%\rstrel{cont}, \rstrel{attr}, \rstrel{evid}, \rstrel{comp},
%\rstrel{list}, \rstrel{back}, \rstrel{same}, \rstrel{topic},
%\rstrel{mann}, \rstrel{summ}, \rstrel{cond}, \rstrel{temp},
%\rstrel{eval}, \rstrel{text}, \rstrel{cause}, \rstrel{prob},
%\rstrel{elab}.

Work on RST parsing has been dominated by the bottom-up paradigm
\cite{hernault2010hilda,feng-hirst-2014-linear,ji-eisenstein-2014-representation,braud-etal-2017-cross-lingual,morey-etal-2017-much,yu2018transition}.
These methods produce very competitive benchmarks, but in practice it is 
not a straightforward	approach (e.g.\ transition-based parser with 
actions prediction steps). Furthermore, bottom-up parsing limits the 
tree construction to local information, and macro context such as global 
structure/topic is prone to be under-utilized. As a result, there
has recently been a move towards top-down approaches \cite{kobayashi2020top,zhang-etal-2020-top}.
%, particularly for
%transition-based parsers, wherein parsing is decomposed into a series of
%actions using a stack/queue.
\begin{figure}
	\centering
	\includegraphics[width=3in]{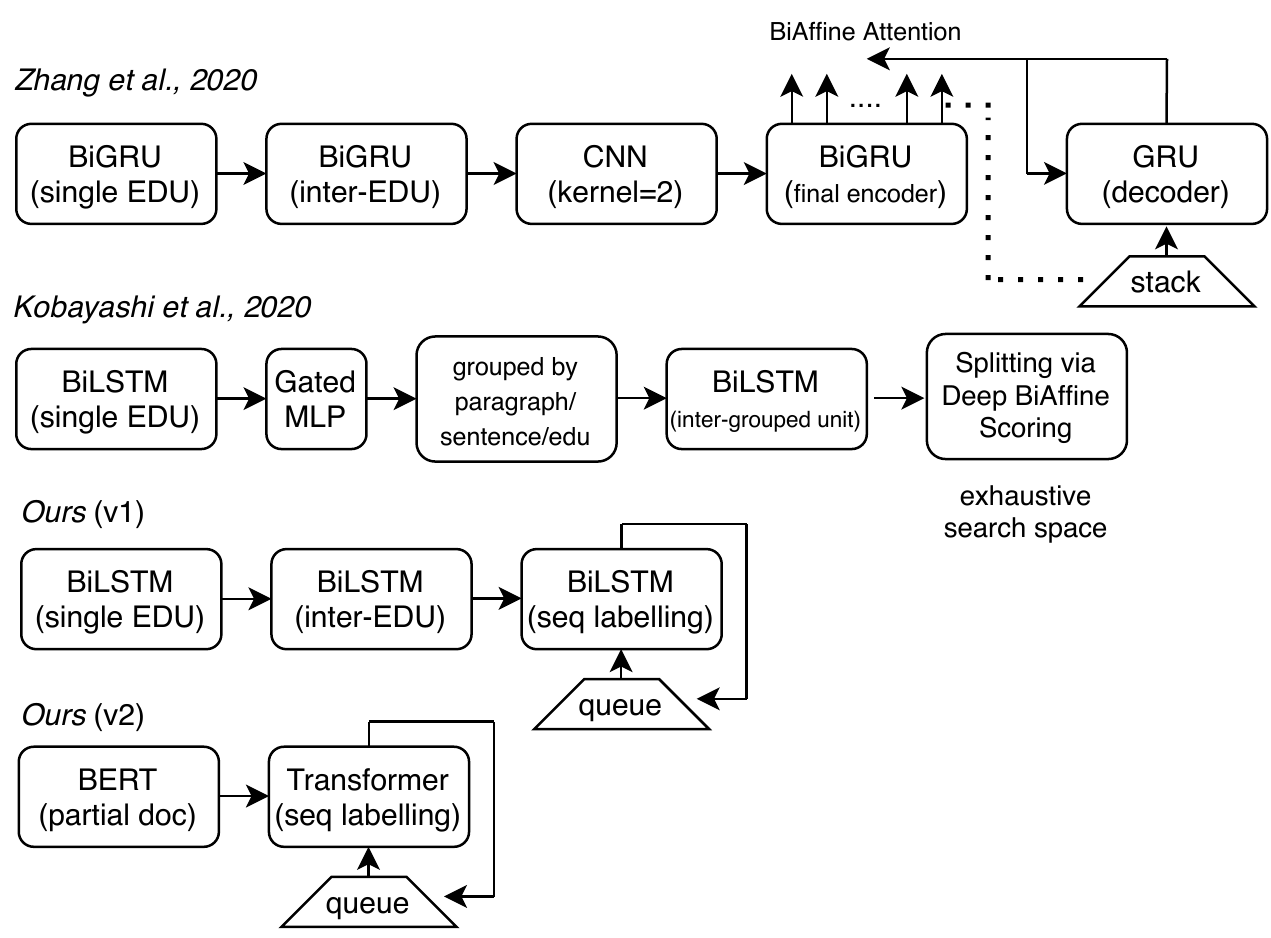}
	\caption{Comparison of our top-down models with
          \citet{zhang-etal-2020-top} and \citet{kobayashi2020top}.}
	\label{fig:compare}
\end{figure}

The general idea behind top-down parsing is to find splitting points in each iteration of tree construction.
In \figref{compare}, we illustrate how our
architecture differs from \citet{zhang-etal-2020-top} and \citet{kobayashi2020top}. First, \citet{zhang-etal-2020-top} utilize four levels of encoder that comprise 3 Bi-GRUs and 1 CNN layer. The splitting mechanism is applied through a decoder, a stack, and bi-affine attention mechanisms.
\citet{kobayashi2020top} use the gold paragraph and sentence boundaries
to aggregate a representation for each unit, and generate the tree based
on these granularities. Two Bi-LSTMs are used, with splitting points
determined by exhaustively calculating the bi-affine score of each
possible split. The use of paragraph boundaries can explicitly lower the
difficulty of the task, as 77\% of paragraphs in the English RST Discourse Treebank (``RST-DT'') are actually text spans \cite{carlson2001building}. These boundaries are closely related to gold span boundaries in evaluation.

In this paper, we propose a conceptually simpler top-down approach for RST parsing. The core idea is to frame the problem as a sequence labelling
task, where the goal is to iteratively find a segmentation boundary to
split a sequence of discourse units into two sub-sequences of discourse
units. This way, we are able to simplify the architecture, in
eliminating the decoder as well as reducing the search space for splitting points. Specifically, we use an LSTM \cite{Hochreiter+:1997}  or pre-trained BERT \cite{devlin-etal-2019-bert} as  the segmenter, enhanced in a number of key ways.

Our primary contributions are as follows: (1) we propose a novel
top-down approach to RST parsing based on sequence labelling; (2) we
explore both traditional sequence models such as LSTMs
and also modern pre-trained encoders such as
BERT; (3) we demonstrate that adding
a weighting mechanism during the splitting of EDU sequences improves 
performance; and (4) we propose a novel dynamic oracle for training 
top-down discourse parsers.

\section{Related Work}

Previous work on RST parsing has been dominated by bottom-up approaches \cite{hernault2010hilda,joty-etal-2013-combining,li-etal-2016-discourse,braud-etal-2017-cross-lingual,wang-etal-2017-two}.
%Traditionally, discourse parsing is carried out via classical machine learning models. For example, \citet{hernault2010hilda} implement a classic, bottom-up greedy approach with a support vector machine (SVM); \citet{joty-etal-2013-combining} apply a two-stage CYK parser with dynamic conditional random fields (CRFs); \citet{feng-hirst-2014-linear} complement a bottom-up greedy approach with linear-chain CRF models; and
For example, \citet{ji-eisenstein-2014-representation} introduce \texttt{DPLP}, a transition-based parser based on an SVM with representation learning,
combined with some heuristic features. \citet{braud-etal-2016-multi}
propose joint text segment representation learning for predicting RST discourse trees using a hierarchical Bi-LSTM. Elsewhere, \citet{yu2018transition} showed that implicit syntax features extracted from a dependency parser \cite{dozat2017deep} are highly effective for discourse parsing.
%, improving results in predicting the EDU span, nuclearity, and relations.

Top-down parsing is well established for constituency parsing
and language modelling \cite{johnson-1995-squibs,roark-johnson-1999-efficient,roark-2001-probabilistic,frost-etal-2007-modular},
but relatively new to discourse parsing. \citet{lin-etal-2019-unified} propose a unified framework based on pointer networks for sentence-level discourse parsing, while \citet{liu2019hierarchical} employ hierarchical pointer network parsers.

%Deep learning models have emerged more recently.
%\citet{braud-etal-2016-multi} propose joint text segment representation for predicting RST discourse trees using a hierarcFhical Bi-LSTM, while
%\citet{li-etal-2016-discourse} employ an attention mechanism with
%a CYK-like parsing process. In other work,
%\citet{braud-etal-2017-cross-lingual} experiment with a transition-based
%discourse parser in a cross-lingual setting.

%: "Full" hasn't been introduced or described; my suggestion is to
%drop the low level discussion of full vs span
\citet{morey-etal-2017-much} found that most previous studies on parsing
RST discourse tree were incorrectly benchmarked, e.g.\ one study uses
macro-averaging while another use micro-averaging.\footnote{After
  standardizing evaluation (based on micro-averaged F-1), they found
  that \texttt{DPLP} achieves the best \texttt{Full} performance,
  outperforming the deep learning models.} They also advocate for
evaluation based on micro-averaged F-1 scores over labelled attachment
decisions (a la the original Parseval).

%With respect to the \texttt{Full} (micro-averaged F-1) metric, the current state-of-the-art bottom-up RST parser is a transition-based neural
%Elsewhere, \citet{yu2018transition} showed that the implicit syntax
%features extracted from a dependency parser \cite{dozat2017deep} are
%highly effective for discourse parsing, improving results in predicting
%the EDU span, nuclearity, and relations.  \citet{wang-etal-2017-two}
%proposed using an SVM for two-stage discourse parsing, which they showed
%to perform marginally better at the \texttt{Span} level than
%\citet{yu2018transition}.

%The notion of top-down discourse parsing was previously proposed by
%\citet{atutxa2019towards} for the Basque language. However, this work
%was limited to the task of detecting the central unit (thesis
%statement), rather than constructing the full discourse tree. In very
%recent work, \citet{kobayashi2020top} apply top-down discourse parsing
%based on different granularity levels. Specifically, they determine
%splitting points by exhaustively calculating the bi-affine score of each
%possible split. As a result, their approach is much slower than ours.

Pre-trained language models
\cite{radford2018building,devlin-etal-2019-bert}
have been shown to benefit a multitude of NLP tasks, including discourse analysis.
%In syntax parsing for example,
%\citet{kitaev-etal-2019-multilingual} demonstrated that BERT is useful
%for multilingual constituency parsing,
%while \citet{mrini2019rethinking}
%improved their dependency parser using BERT and XLNet \cite{yang2019xlnet}.
For example, BERT models have been used for classifying discourse markers
\cite{sileo-etal-2019-mining} and discourse relations
\cite{nie-etal-2019-dissent,shi2019next}. To the best of our knowledge,
however, pre-trained models have not been applied in the generation of
full discourse trees, which we address here by experimenting with BERT
for top-down RST parsing.

\section{Top-down RST Parsing}

%Several RST parsers has been modelled with neural architecture to
%obtain EDU representation \cite{li-etal-2014-recursive,
%li-etal-2016-discourse,yu2018transition}. However, none of them has
%shown how pre-trained-encoder such as BERT \cite{devlin-etal-2019-bert}
%contributes to discourse tree construction. Therefore we divide this
%work into two different encoders: 1) standard RNN with LSTM cell and 2)
%Transformer with its powerful BERT.

%JHL: technically EDU is the "elementary" discourse unit, which are the
%leaves; the intermediate nodes in the tree are discourse units (DUs).
%We can potentially fix this and generalise everything by calling them
%DUs, or just leave the text as they are and call all the DUs EDUs, up
%to you Fajri
We frame RST parsing as a sequence labelling task, where given a
sequence of input EDUs, the goal is to find a segmentation boundary to
split the sequence into two sub-sequences. This is realized by training
a sequence labelling model to predict a binary label for each EDU, and
select the EDU with the highest probability to be the segmentation
point.  After the sequence is segmented, we repeat the same process for
the two sub-sequences in a divide-and-conquer fashion, until all sequences
are segmented into individual units, producing the binary RST tree
(e.g.\ \figref{rsttree}).

%For the sequence labelling models, we experiment with an LSTM
%\cite{Hochreiter+:1997} and BERT \cite{devlin-etal-2019-bert}.

\subsection{LSTM Model}
\label{sec:lstm}

\begin{figure}
	\centering
	\includegraphics[width=3.0in]{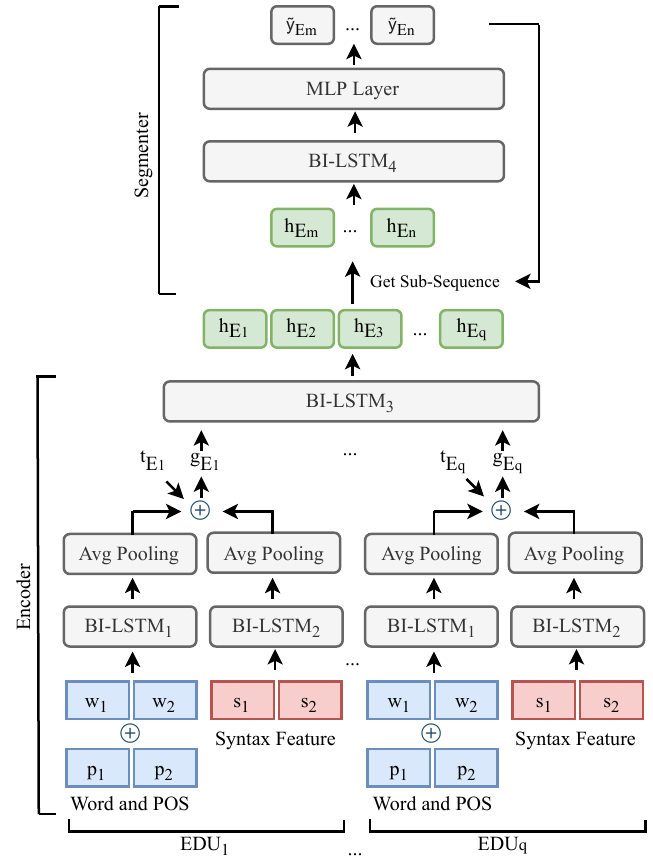}
    \caption{Architecture of the LSTM model.}
	\label{fig:rstrnn}
\end{figure}

As illustrated in \figref{rstrnn}, our LSTM parser consists of two main
blocks: an encoder and a segmenter. For the encoder, we follow
\citet{yu2018transition} in using two LSTMs (Bi-LSTM$_1$ and
Bi-LSTM$_2$) to produce EDU encodings by processing: (1) $x_i$, the
concatenation of word embedding $w_i$ and POS tag embedding $p_i$; and
(2) syntax embedding $s_i$,  the output of the MLP layer of
the bi-affine dependency parser \cite{dozat2017deep}. Similar to \citet{yu2018transition}, we then take the
average of the output states for both LSTMs over the EDU, and
concatenate it with an EDU type embedding $t_{E_j}$ (which distinguishes
the last EDU in a paragraph from other EDUs) to produce the final
encoding:
\begin{align}
    x_i &= w_i \oplus p_i \nonumber \\
\{a_1^w, .., a_p^w\} &= \text{Bi-LSTM}_1(\{x_1,.., x_p\}) \nonumber \\
\{a_1^s, ..., a_p^s\} &= \text{Bi-LSTM}_2(\{s_1,.., s_p\}) \nonumber \\
g_{E_j} &= \text{Avg-Pool}(\{a_1^w,.., a_p^w\}) \oplus \nonumber \\
&\phantom{= } \text{Avg-Pool}(\{a_1^s,.., a_p^s\}) \oplus t_{E_j}
\label{eqn:lstm-syntax}
\end{align}
where $E_j$ is an EDU, $p$ is the number of words in $E_j$, and $\oplus$
denotes the concatenate operation. $t_{E_j}$ is generally an implicit paragraph boundary feature, and provides a fair benchmark with previous models. In \secref{result}, we also show results without paragraph boundary features.

%\footnote{In \secref{result}, we also show result without paragraph boundary features. We include this feature to allow us doing a fair benchmark with some previous models.}
%JHL: what is this EDU type embedding? what is middle or end of
%paragraph? whether the EDU is in the middle of a paragraph or the last
%EDU in a paragraph? This is all very strange
%FJ: it means to distinguish the last EDU in a paragraph with other EDUs

As each EDU is processed independently, we use another LSTM
(Bi-LSTM$_3$) to capture the inter-EDU relationship to obtain a
contextualized representation $h_{E_j}$:
\begin{align*}
\{h_{E_1}, ..., h_{E_q}\} &= \text{Bi-LSTM}_3(\{g_{E_1}, ..., g_{E_q}\})
\end{align*}
where $q$ is the number of EDUs in the document. Note that $h_{E_j}$ is the final encoder output (see \figref{rstrnn}) and is only computed once for each document.

The second part is the segmenter. We frame segmentation as a
sequence labelling problem with $y_{E_j} \in \{0, 1\}$, where 1 denotes
the splitting point, and 0 a non-splitting point. For each EDU sequence there is
exactly one EDU that is labeled 1, and we start from the full EDU
sequence (whole document) and iteratively perform segmentation until we
are left with individual EDUs.  We use a queue to store the two
EDU sub-sequences as the result of the segmentation process. In total,
there are $q-1$ iterations of segmentation (recall that $q$ is the total
number of EDUs in the document).

As segmentation is done iteratively in a divide-and-conquer fashion,
$h_{E_j}$ serves as the input to the segmenter, which takes a
(sub)sequence of EDUs to predict the segmentation position:
\begin{align*}
\{h'_{E_m},..,h'_{E_n}\} &= \text{Bi-LSTM}_4(\{h_{E_m},..,h_{E_n}\})\\
\tilde{y}_{E_j} &= \sigma(\text{MLP}(h'_{E_j}))
\end{align*}
where $m$/$n$ are the starting/ending index of the EDU
sequence,\footnote{In the first iteration, $m=1$ and $n=q$ (number of
EDUs in the document).} and $\tilde{y}_{E_j}$ gives the probability of a
segmentation. From preliminary experiments we found that it's important
to have this additional Bi-LSTM$_4$ to perform the EDU sub-sequence
segmentation point prediction.

\subsection{Transformer Model}
\label{sec:transformer}

%Although pre-trained models such as BERT have been shown to perform well
%over sequence labelling tasks,
Adapting BERT to discourse parsing is not
trivial due to the limited number of input tokens it takes (typically
512 tokens), which is often too short for documents. Moreover, BERT is
designed to encode sentences (and only two at maximum), where in our
case we want to encode sequences of EDUs that span multiple
sentences.

In our case, EDU truncation is not an option (since that would produce an
incomplete RST tree), and %as \tabref{datastat} shows,
the average number  of words per document in our data is 521 (741 word pieces after BERT tokenization), which is much larger than the 512 limit. We therefore break the document into a number of partial documents, each consisting of multiple sentences that fit into the 512 token limit. This way, we allow the model to capture the fine-grained
word-to-word relationships across (most) EDUs. Each partial document is then processed 
based on \citet{liu2019text} trick where we use an alternating even/odd segmentation embedding to encode all the EDUs in a document.

\begin{figure}
	\centering
	\includegraphics[width=3.1in]{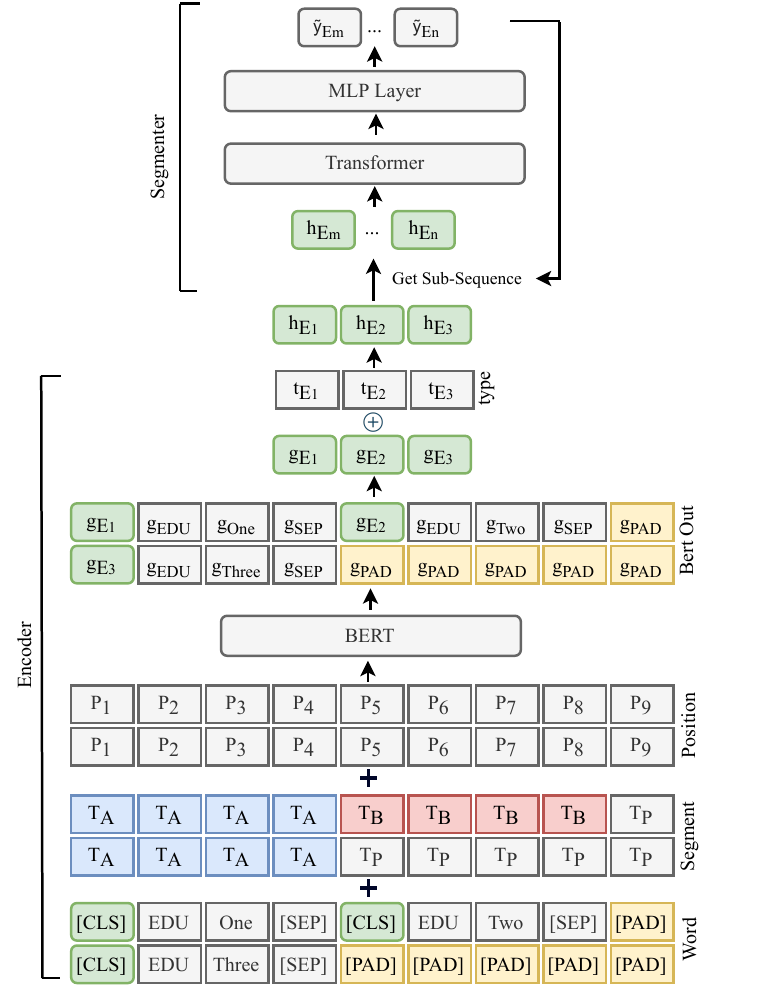}
    \caption{Architecture of the transformer model. In practice, 1 row of input can have more than two EDUs. }
    \label{fig:rstbert}
\end{figure}

We illustrate this approach in \figref{rstbert}. First, all EDUs are
formatted to start with [CLS] and end with [SEP], and words are
tokenized using WordPiece. If the document has more than 512 tokens, we
break it into multiple partial documents based on EDU boundaries,
and pad accordingly (e.g.\ in \figref{rstbert} we break the example
document of 3 EDUs into 2 partial documents), and process each partial
document independently with BERT.

We also experimented with the second alternative by 
encoding each EDU independently first with BERT, and use a
second inter-EDU transformer to capture the relationships between EDUs.
Preliminary experiments, however, suggest that this approach produces
sub-optimal performance.

%\citet{liu2019text} address these issues by introducing
%additional positional embeddings (for inputs with more than 512 tokens)
%and using an alternating even/odd segmentation embedding to encode all
%the sentences in a document for summarization. In practice, however,
%\citet{liu2019text} limit the document length to
%512 tokens, as it was demonstrated to be empirically effective. In our
%case, EDU truncation is not an option (since that would produce an
%incomplete RST tree), and 
%the average number  of words per document in our data is 521 (741 word pieces after BERT tokenization), which is much larger than the 512 limit.
%even limiting the token length of each EDU to 20 words, for instance,
%does not result in less than 512 tokens for all documents (see
%\tabref{datastat} for \#word distributions).

%We experiment with two ways to tackle this problem. First, similarly to our LSTM encoder, we can encode each EDU independently first with BERT, and use a
%second inter-EDU transformer to capture the relationships between EDUs.
%Preliminary experiments, however, found that this approach produces
%sub-optimal performance.

%  However, there is
%no guarantee of an optimal result
%as this involves discarding segment information. Alternatively, we
%can reduce available EDUs down to 512 tokens, and shard the input if it is
%greater than 512.
%% TJB: need to say something about sharding, or add reference
%In this way, we are able to fully harness BERT to
%generate RST parses.

% use sharding if a document has more than 512 tokens, and otherwise pad
% to make up 512 tokens. Our method for segmenting EDUs is based on
% \citet{liu2019text}.

In \figref{rstbert} each token is assigned three kinds of embeddings: (1) word, (2) segment,
and (3) position. The input vector is computed by summing these three
embeddings, and fed into BERT (initialized with \texttt{bert-base}).
The output of BERT gives us a contextualized embedding for each token,
and we use the [CLS] embedding as the encoding for each EDU ($g_{E_j}$).
%\begin{align}
%\begin{split}
%\hat{h}^l &= \text{LN}(h^{l-1} + \text{MHAtt}(h^{l-1})) \\
%h^l &= \text{LN}(\hat{h}^l + \text{FFN}(\hat{h}^l))
%\end{split}
%\label{eqn:transformer}
%\end{align}
%where $h^0$ is the input vectors $x$, $l$ denotes the stacked layers of
%the transformer, $\text{LN}$ is layer normalisation, and $\text{MHAtt}$
%is the multi-headed attention function
%\cite{vaswani2017attention}.

Unlike the LSTM model, we do not incorporate syntax embeddings into the
transformer model as we found no empirical benefit (see
\secref{result}).  This observation is in line with other studies (e.g.\
\citet{jawahar-etal-2019-bert}) that have found BERT to implicit encode
syntactic knowledge.

%To incorporate the syntax features ($s_i$), we first average them over
%the EDU, compress them with a linear layer and finally concatenate it
%with the EDU encoding $g_{E_j}$:
%\begin{align}
%s'_{E_j} &= \text{Avg-Pool}(\{s_1, ..., s_p\}) \nonumber \\
%\hat{s}_{E_j} &= W_s s'_{E_j} + b_s \label{eqn:compress} \\
%h_{E_j} &= g_{E_j} \oplus \hat{s}_{E_j} \label{eqn:transformer-syntax}
%\end{align}
%where AvgEDU is average pooling over syntax vectors in an EDU, and
%$t_{E_i}$ is the encoded [CLS] of $\text{EDU}_i$. Unlike the RNN
%architecture in Figure~\ref{fig:rstrnn}, we use only BERT as the encoder
%because the inter-EDU relationship has been partially learned via the shards.

For the segmenter we use a second transformer (initialized with random
weights) to capture the inter-EDU relationships for sub-sequences of
EDUs during iterative segmentation:
\begin{align*}
\{h'_{E_m},..,h'_{E_n}\} &= \text{transformer}(\{h_{E_m},..,h_{E_n}\})\\
\tilde{y}_{E_j} &= \sigma(\text{MLP}(h'_{E_j}))
\end{align*}
where $\tilde{y}_{E_j}$ gives the probability of a segmentation, and
 $h_{E_j}$ is the concatenation of the output of BERT ($g_{E_j}$) and
the EDU type embedding ($t_{E_j}$).

% Data distribution
%\begin{table}[t]
%	\footnotesize
%	\begin{center}
%		\begin{adjustbox}{max width=\linewidth}
%			\begin{tabular}{lccc}
%				\toprule
%				\textbf{Stats} & \textbf{Mean} & \textbf{SD} & \textbf{Max} \\
%				\midrule
%				EDUs/Doc & 56.0 & 52.5 & 304 \\
%				% \multicolumn{4}{l}{\textit{Original Document}} \\
%				Words/EDU & 9.46 & 6.05 & 60 \\
%				Words/Doc & 529 & 470 & 2607 \\
%				% \midrule
%				%Tokens/EDU & 13.3 & 7.5 & 98 \\
%				%1Tokens/Doc & 741 & 661 & 3615 \\
%				\bottomrule
%			\end{tabular}
%		\end{adjustbox}
%	\end{center}
%	\caption{Dataset statistics.}
%	\label{tab:datastat}
%\end{table}

\subsection{Nuclearity and Discourse Relation Prediction}

\begin{figure}[]
	\centering
	\includegraphics[width=3in]{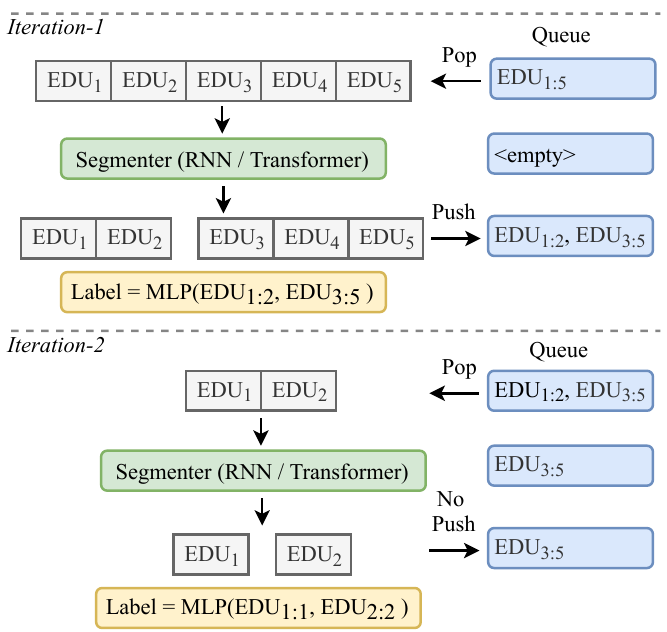}
	\caption{Nuclearity and relation prediction.}
	\label{fig:topdown}
\end{figure}

In \figref{topdown}, we give an example of the iterative segmentation
process to construct the RST tree. In each iteration, we pop a sequence
from the queue (initialized with the original sequence of EDUs in the
document) and compute the segmentation label for each EDU using an LSTM
(\secref{lstm}) or transformer (\secref{transformer}). After the
sequence is segmented (using the ground truth label during training, or
the highest-probability label at test time), we push to the queue the
two sub-sequences (if they contain at least two EDUs) and repeat this
process until the queue is empty.

%we apply a
%divide-and-conquer scheme by tagging EDUs in a sequence with labels 0 or
%1, where 1 indicates the segmentation position. Because there exists
%exactly one segmentation position within a given sequence, there is only
%one EDU with label 1. Subsequently, by using the segmentation position,
%we divide the sequence into two sub-sequences and we push them to the
%queue if it contains at least two EDUs. The process is terminated when
%there is no element left in the queue.

In addition to segmentation, we also need to predict the
nuclearity/satellite relationship (3 classes) and the discourse label
(18 classes) for the segmented pairs. To that end, we average the EDU
encodings for the segments, and feed them to a MLP layer to predict the
nuclearity and discourse labels:
\begin{align*}
u_{l} &= \text{Avg-Pool}(h'_{E_m}, ... , h'_{E_{m+ind}})  \\
u_{r} &= \text{Avg-Pool}(h'_{E_{m+ind+1}}, ..., h'_{E_{n}}) \\
z_{nuc+dis} &= \text{softmax}(\text{MLP}(u_l, u_r))
\end{align*}
where $ind$ is the index of the segmentation point (given by the ground
truth during training, or argmax of the segmentation probabilities
$\tilde{y}_{E_j}$ at test time), and ${z}_{nuc+dis}$ gives the joint
probability distribution over the nuclearity and discourse classes.\footnote{We also experimented with predicting the nuclearity and discourse labels separately, but found joint prediction to work better in preliminary experiments.}

%We predict nuclearity and the relation type with two different MLPs
%as shown above, or use one MLP layer by combining the labels. This
%minimal difference has significant impact on the top-down approach, and will
%be discussed further in \secref{experiments}.

%As nuclearity and discourse labels are correlated, we also
%experiment with an approach that predicts both labels jointly (the total
%number of classes for the prediction is $ 3 \times 18 = 54$ in theory,
%but in practice there are only 41 classes, as certain combinations of
%nuclearity and relations are not attested in the training data):

\subsection{Segmentation Loss with Penalty}
\label{sec:penalty}

One drawback of the top-down approach is that segmentation
errors incurred closer to the root can be detrimental, as the error will
propagate to the rest of the sub-trees. To address this, we explore
scaling the segmentation loss based on the current tree depth and the
number of EDUs in the input sequence. Preliminary experiments found that
both approaches work, but that the latter is marginally better, and so we
present results using the latter.

%penalties.  The first is via the \textit{depth} of the tree, based on
%the
%intuition that shallower nodes are critical to segmentation. The second
%is via the \textit{number of elements} in the sequence, where the more
%EDUs in a sequence, the riskier the segmentation.
% TJB: not sure I understand this last sentence: use them for what?

%In this work, we only apply \textit{number of elements} as an
%additional penalty, because preliminary experiments using \textit{depth} did not
%show any improvement. % Although there might be a correlation between
% \textit{depth} and \textit{number of element}, please keep in mind that
% in the lower depth (early segmentation), we might have sequences with
% few EDUs as the result of segmentation. This implies that using
% \textit{number of element} as the penalty is intuitively more accurate.

Formally, the modified segmentation loss of an example (document) is
given as follows:
\begin{align*}
L(E_{m:n}) &= - \sum_{i=m}^{n} \Big(y_{E_i} \log(\tilde{y}_{E_i}) + \\
    &\phantom{=} (1-y_{E_i}) \log(1 - \tilde{y}_{E_i}) \Big) \\
\mathcal{L}_{seg} &= \frac{1}{|S|} \sum_{(m,n) \in S} (1 + (n-m)^\beta)
  L(E_{m:n})
\end{align*}
where $y_{E_i} \in \{0, 1\}$ is the ground truth segmentation label,
$L(E_{m:n})$ is the cross-entropy loss for an EDU sequence, $S$ is the
set of all EDU sequences (based on ground truth segmentation), and
$\beta$ is a scaling hyper-parameter.

To summarize, the total training loss of our model is a (weighted)
combination of segmentation loss  ($\mathcal{L}_{seg}$) and
nuclearity-discourse prediction loss
($\mathcal{L}_{nuc+dis})$:
\begin{equation}
\mathcal{L} = \lambda_1 \mathcal{L}_{seg} + \lambda_2 \mathcal{L}_{nuc+dis}
\label{eqn:separate-loss}
\end{equation}

\subsection{Dynamic Oracle}
\label{sec:dynamic-oracle}

The training regimen for discourse parsing creates an exposure bias, where
the parser may struggle to recover when it makes a mistake at test time.
\citet{goldberg-nivre-2012-dynamic} propose a dynamic oracle for
transition-based dependency parsing to tackle this. The idea is to allow
the model during training to use its predictions (instead of ground
truth actions), and introduce a dynamic oracle to find the next
best/optimal action sequences.  It does so by comparing the current
state of the constructed tree and the gold-standard tree, and aims to
minimize the deviation. As the model is exposed to prediction errors
during training time, it has a better chance of recovering from them at
test time.

 \begin{algorithm}[t]
    \small
	\caption{Top-down Dynamic Oracle}
        \label{alg:oracle}
	\begin{algorithmic}[1]
		\Function{DynOracle}{$E,O, R$}
		\State{\# \textit{For training only}}
		\State{\# \textit{E is list of EDUs}}
		\State{\# \textit{O is gold order for segmentation}}
		\State{\# \textit{R is list of gold discourse labels based on O}}
		\State $q = $ length$(E)$; $queue = [E_{1:q}]$
		\While{$queue$ is not empty}
			\State $E_{m:n} = queue$.pop()
			\State $id_{gold}, r_{gold}$ = match$(E_{m:n}, O, R)$
			\State $id_{pred} =\text{predictSplit}(E_{m:n})$
			\State $r_{pred1} =\text{predictLabel}(E_{m:n}, id_{gold})$ \# \textit{for loss}
			\State $r_{pred2} =\text{predictLabel}(E_{m:n}, id_{pred})$ \# \textit{ignored}
			\If {random$()>\alpha$}
				\State $L,R=\text{separate}(E_{m:n},id_{gold})$
			\Else
				\State $L,R=\text{separate}(E_{m:n},id_{pred})$
			\EndIf
			\State $queue$.push$(L)$ \textbf{if} len$(L)>1$
			\State $queue$.push$(R)$ \textbf{if} len$(R)>1$
		\EndWhile
		\EndFunction
	\end{algorithmic}
\end{algorithm}

% Intuitively, incorporating optimal actions during the training will
%guide the model to produce the best possible tree as the error at a
%testing time can not be avoided.  TJB: description needs tightening up
% FJ: Updated

We explore a similar idea, and propose a dynamic oracle for our top-down
discourse parser.
A crucial question to ask when designing a dynamic oracle is:
\textit{how can we compare the current state to the gold tree to obtain
the next best series of actions when an error occurs during training?}
In transition-based parsing, \citet{goldberg-nivre-2012-dynamic} compute
a cost/loss of each transition by counting the gold arcs that are no
longer reachable based on the action taken (e.g. \texttt{SHIFT},
\texttt{REDUCE}). We apply similar reasoning when finding the next
best segmentation sequence in our dynamic oracle, which we illustrate
below with an example.

% Following this, our dynamic oracle for top-down parsing works as
%follows.

%Let $J$ be a list of EDU sequence, where $|J|=q-1$ is the total number
%of steps (splits) required to construct a full tree, and $q$ is the
%number of EDU in the document. $J_i$ is an EDU sequence $E_{m:n} =
%[E_m, E_{m+1}, .., E_n]$ with $m<n$. For each $J_i$ there exists a
%binary sequence\footnote{a sequence that exactly has one  label 1, and
%0 for the rest.} $K_i$ denoting the gold segmentation label for $J_i$.
%Thus, in this case $|K_i| = |J_i| = n-m+1$, and $|K| = q-1$.  Because
%there is only $n-m$ possible splitting position for $E_{m:n}$, we can
%ignore the last element in $K_i$ that denotes label for $E_n$. In this
%case, if the first element in $K_i$ is 1 then the split will be applied
%between $E_m$ and $E_{m+1}$.

\begin{figure}
	\centering
	\includegraphics[width=1\linewidth]{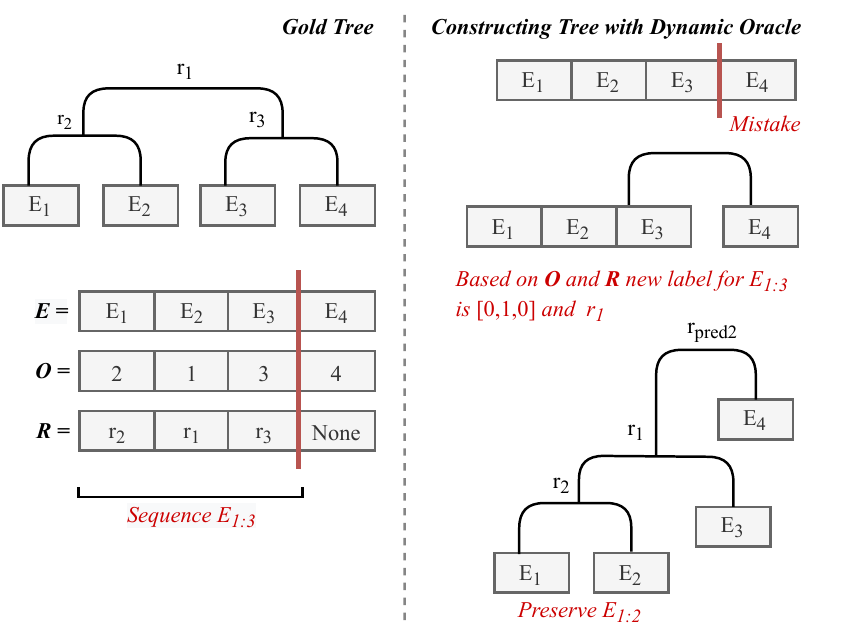}
	\caption{Dynamic oracle for top-down approach.}
	\label{fig:dynamic}
\end{figure}

%Now given $J$ and $K$, we can obtain a gold order $O$ with length $q$,
%an integer list indicating the splitting order to construct a full gold
%tree. The last index in $O$ can be ignored as there is only $q-1$
%possible splits. As an example,
Say we have a document with 4 EDUs ($E_{1:4}$), and the gold tree given
in \figref{dynamic} (left). The correct sequence of segmentation is
given by $O_{1:4}=[2,1,3,-]$, which means we should first split at $E_2$
(creating $E_{1:2}$ and $E_{3:4}$), and then at $E_1$ (creating
$E_{1}, E_{2}, E_{3:4}$), and lastly at $E_3$, producing
$E_1, E_2, E_3, E_4$ as the leaves with the gold tree structure. We give
the last EDU $E_4$ a ``$-$'' label (i.e.\ $O_4 = $`$-$') because no
segmentation is needed for the last EDU.

Suppose the model predicts to do the first segmentation at $E_3$. This
produces $E_{1:3}$ and $E_4$. What is the best way to segment $E_{1:3}$
to produce a tree that is as close as possible to the gold tree? The
canonical segmentation order $O_{1:3}$ is $[2,1,-]$ (the label of the
last EDU is replaced by `$-$'), from which we can see the next best segmentation is
to segment at $E_2$ to create $E_{1:2}$ and $E_3$. Creating the
canonical segmentation order $O$, and following it as much as possible,
ensures the sub-tree that we're creating for $E_{1:3}$ mimics the
structure of the gold tree.

% initially we have the full EDU sequence $E_{1:4}=[E_1, E_2, E_3,
% E_4]$.
%Suppose we have $O_{1:4}=[2,1,3,4]$ as the gold segmentation order,
%then  the first split will be applied between $E_2$ and $E_3$ resulting
%in $E_{1:2}$ and $E_{3:4}$. The second split is between $E_1$ and
%$E_2$, and the third one is between $E_3$ and $E_4$.

%By having $O$ we can apply dynamic oracle for top-down parsing.  We
%describe this approach in Figure~\ref{fig:dynamic} and
%\algoref{oracle}.  The idea is to match any given sequence $E_{m:n}$
%with order $O_{m:n}$ and set the gold splitting point by the earliest
%order in $O_{m:n}$. For instance, for $E_{1:3}$ the order $O_{1:3}$ is
%$[2,1,3]$, thus the split will be applied between $E_2$ and $E_3$. For
%$E_{1:2}$ the order $O_{1:2}$ is $[2,1]$ and the split is applied
%between $E_1$ and $E_2$ as we always ignore the last index at $O$.

%JHL: update figure to use E_1 instead of EDU_1 (consistent with text)
%FJ: done
The dynamic oracle labels nuclearity-discourse relations following the
same idea.  We introduce $R$, a list of gold nuclearity-discourse
relations.  For our example $R_{1:4}=[r_2, r_1, r_3, -]$ (based on the
gold tree; see \figref{dynamic} (left)).  If the model decides to first
segment at $E_3$ and creates $E_{1:3}$ and $E_4$, when we segment at
$E_2$ (next best choice of segmentation), we will follow $R$ and label
the nuclearity-discourse relation with $r_1$.  As before, following the
original label list $R$ ensures we keep the nuclearity-discourse relation
as faithful as possible (\figref{dynamic} (right bottom)).

% before, for $E_{1:3}$ we pick the earliest order from $R_{1:3}$, which
%is $r_1$ as the relation.  This way, we are partially able to recover
%the tree from its error state.

%Nuclearity and discourse labels for unattested EDU sequences (i.e.\
%sequences that are not seen in the ground truth) can be determined by
%following the original order of segmentation.  For instance in
%\figref{dynamic}, as the input sequence EDU$_{1:4}$ (created in the
%first iteration) does not exist in the ground truth, and the best
%discourse label for EDU$_{1:2}$ and EDU$_{3:4}$ is $r_1$.
% By looking at the earliest segmentation order within an EDU
%sub-sequence (see Figure~\ref{fig:dynamic}), we can match any sequence
%with its new labels.

%When evaluating the nuclearity and discourse labels, however, it is
%dependent on the correct prediction of the input EDU sequence. For the
%example in \figref{dynamic}, the discourse label for the input sequence
%EDU$_{1:4}$ will always be judged to be wrong as the span is unattested
%in the ground truth in the first place.  Following this, we experiment
% with an approach where we implement the dynamic oracle only for segmentation,
% while for discourse labels they are learnt following the ground truth %segmentations. The
%drawback, of course, is that training becomes more expensive as we need
%to track two different routes for back-propagation. For this reason, we
%only test this for the LSTM model (\secref{lstm}).

The dynamic oracle of our top-down parser is arguably quicker than that
of a transition-based parser, as we do not need to accumulate cost for
every transition taken.  Instead, the dynamic oracle simply follows the
gold segmentation order $O$ to preserve as many subtrees as possible
when an error occurs. We present pseudocode for the proposed dynamic
oracle in \algoref{oracle}.

The probability of using the ground truth segmentation or predicted
segmentation during training is controlled by the hyper-parameter
$\alpha\in[0,1]$ (see \algoref{oracle}). Intuitively, this hyper-parameter
allows the model to alternate between exploring its (possibly erroneous)
segmentation or learning from the ground truth segmentation. The oracle
reverts to its static variant when $\alpha=0$.

\section{Experiments}
\label{sec:experiments}

\subsection{Data}

We use the English RST Discourse Treebank \cite{carlson2001building} for our
experiments, consistent with recent studies
\cite{ji-eisenstein-2014-representation, li-etal-2014-recursive,
  feng-hirst-2014-linear, yu2018transition}. The dataset is based on the
Wall Street Journal portion of the Penn Treebank
\citep{marcus-etal-1993-building}, with 347 documents for
training, and the remaining 38 documents for testing. We use the same
development set as \citet{yu2018transition}, which consists of 35
documents selected from the training set. We also use
the same 18 discourse labels. Stanford
CoreNLP \cite{manning-etal-2014-stanford} is used for POS
tagging.\footnote{\url{https://stanfordnlp.github.io/CoreNLP}}
% TJB: do we really lemmatise? is this a good idea given that we are
% using BERT?
% FJ: No its the mistake, I have checked the document and they are still complete.

%We provide statistics of the dataset in \tabref{datastat}. On average
%there are 56 EDUs in each document, and almost 10 words per EDU.

\subsection{Model Configurations}

We experiment with two segmentation models --- LSTM (\secref{lstm}) and
transformer (\secref{transformer}) --- both implemented in PyTorch
framework.\footnote{We use the Huggingface framework for the transformer
  models.} As EDUs are provided in the dataset, no automatic
segmentation of EDU is required in our experiments.

For the LSTM model, the dimensionality of the Bi-LSTMs in the encoder is
256, while the segmenter (Bi-LSTM$_4$) is 128
(\figref{rstrnn}).
% TJB: do you mean 50? The max length is 60, right?
% FJ: I will delete it. Apparently we use all words in LSTM
The embedding dimensions of words,
POS tags, EDU type, and syntax features are 200, 200, 100, and 1,200, respectively, and we initialize words in EDU with GloVe embedding
\cite{pennington-etal-2014-glove}.\footnote{\url{https://nlp.stanford.edu/projects/glove}}   For hyper-parameters, we use the following:
batch size $=$ 4, gradient accumulation $=$ 2, learning rate $=$ 0.001,
dropout probability $=$ 0.5, and optimizer = Adam (with epsilon of 1e-6). The loss
scaling hyper-parameters (\eqnref{separate-loss}), are tuned based
on the development set, and set to $\lambda_1 = 1.0$, and $\lambda_2 =
1.0$.

\begin{table}[t]
	\begin{center}
		\begin{adjustbox}{max width=0.9\linewidth}
			\begin{tabular}{lcc}
				\toprule
				\textbf{Variant} &  \textbf{LSTM} &
				\textbf{Transformer} \\
				\midrule
				Vanilla &    48.4$\pm$0.5 & 51.3$\pm$0.2 \\
				$+$Syntax &  50.0$\pm$0.7 & 51.9$\pm$0.4  \\
				$+$Penalty & 49.6$\pm$0.5 & \textbf{52.1$\pm$0.4}\\
				$+$Syntax$+$Penalty &  \textbf{51.6$\pm$0.1} & 51.8$\pm$0.8 \\
				\bottomrule

			\end{tabular}
		\end{adjustbox}
	\end{center}
	\caption{Feature addition study over the development set to find the best configuration for our models. Presented results are the mean and standard deviation of the \texttt{Full} metric (micro-averaged F-score on labelled attachment decisions) over three runs. }
	\label{tab:analysis}
\end{table}
%\begin{figure}
%	\centering
%	\includegraphics[width=3in]{image/elempenalty.pdf}
%	\caption{Effect of $\beta$ in applying element penalty.}
%	\label{fig:elem}
%\end{figure}
%
%\begin{figure}
%	\centering
%	\includegraphics[width=3in]{image/effectdyn.pdf}
%	\caption{Effect of $\alpha$ in applying dynamic oracle.}
%	\label{fig:oracle}
%\end{figure}

%To reduce memory consumption we compress the dimensionality of the
%syntax features from 1,200 to 320 with a linear layer (\eqnref{compress}).

For the transformer model, the document length limit is set to 512
tokens, and longer documents are broken into smaller partial documents.
As before, we truncate each EDU to the first 50 words.
We initialize the transformer in the encoder with \texttt{bert-base},
and the transformer in the segmenter with random weights
(\figref{rstbert}).  The transformer segmenter has 2 layers with 8 heads
and 2048 feed-forward hidden size.  The training hyper-parameters are: initial learning rate $=$ 5e-5, maximum epochs $=$ 250,
warm up $=$ 2000 steps, and drop out $=$ 0.2.  For the $\lambda$
hyper-parameters, we use the same configuration as for the LSTM
model.

We tuned the segmentation loss penalty hyper-parameter $\beta$
(\secref{penalty}) and the dynamic oracle hyper-parameter $\alpha$
(\secref{dynamic-oracle})  based on the development set.
%In Figure~\ref{fig:elem} we found that $\beta$ in between 0.2 and
%0.4 gives higher F1 score for metric \texttt{Full}, while in
%  Figure~\ref{fig:oracle}, the suitable $\alpha$ for dynamic oracle is
%greater than 0.
Both the LSTM and transformer models use the same $\beta=0.35$ and
$\alpha=0.65$. We activate the dynamic oracle after training for 50
epochs for both models.
% In this experiment, we do not use beam search as we found it only
%effective to improve the metric  \texttt{Span} at epoch 0, but not
%provide significant contribution after the model converges.

In terms of evaluation, we use the standard metrics introduced by
\citet{marcu2000the}: \texttt{Span}, \texttt{Nuclearity},
\texttt{Relation}, and \texttt{Full}. We report micro-averaged F-1 scores on
labelled attachment decisions (original Parseval), following the 
recommendation of \citet{morey-etal-2017-much}.  Additionally, we also 
present the evaluation with RST-Parseval procedure in 
\appref{sec:appendix_res}.
%\appref{sec:appendix_res}
%a more accurate evaluation of RST parser performance.

% For most experiments, we utilize NVIDIA V100 where we use 1 GPU and 4
%GPUs for LSTM and BERT respectively.

%\begin{figure}
%	\centering
%	\includegraphics[width=3in]{image/beam-span.pdf}
%	\caption{Experiment using different beam for segmentation.}
%	\label{fig:beam}
%\end{figure}

\subsection{Results}
\label{sec:result}

We first perform a feature addition study over our models to find the best model
configuration; results are presented in \tabref{analysis}. Note that
these results are computed over the \textit{development set}, based on a static oracle.

For the vanilla models, the transformer model performs much better than
the LSTM model.  Adding syntax features ($+$Syntax) improves both models, although
it's more beneficial for the LSTM. A similar trend is observed when we
modify the segmentation loss to penalize the model if a segmentation
error is made with more EDUs in the input sequence ($+$Penalty;
\secref{penalty}): the transformer model sees an improvement of $+0.8$ while the
LSTM model improves by $+1.2$.  Lastly, when we combine both syntax features and
the segmentation penalty, the LSTM model again shows an appreciable improvement, while
the transformer model drops in performance marginally.\footnote{The result is consistent with the test set (see \appref{sec:appendix_res2})} Given these results, we
use both syntax features and the segmentation penalty for the LSTM model, but only the
segmentation penalty for the transformer model in the remainder of our experiments.

%\appref{sec:appendix_res2}
% in \tabref{analysis} is the vanilla models of top-down approach
% surpass the transition-based model. Particularly, BERT shows its
% primacy in the vanilla models with $60.3$ F1-score for \texttt{Full}
% ($+2.0$ gap with the baseline). Further, the syntax features
% ($+$Syntax) introduced by \citet{yu2018transition} impact on our both
% models, although the improvement is not as substantial as in
% transition-based model.

%Introducing segmentation loss ($+$Penalty) for top-down approach also
%enhances the \texttt{Full} scores of both models, and combining both
%syntax and penalty affects the proposed models in different ways. Our
%LSTM model improves significantly, reaching F-1 score $61.3$ with
%$+1.9$ gap with the vanilla. In BERT model, the best configuration
%appears with only using the penalty. Arguably, as BERT learns syntax
%representations \cite{jawahar-etal-2019-bert}, the improvement with
%these attributes is not compelling. Also, the token limitation of BERT
%leave a room for future work, especially in designing a better
%transformer-based model for long document.

We next benchmark our models against state-of-the-art RST parsers over
the \textit{test set}, as presented in \tabref{mainresult} (original Parseval) and \tabref{mainresult2} (RST-Parseval as additional result).  Except \citet{yu2018transition}, all bottom-up results are from \citet{morey-etal-2017-much}. We present the labelled attachment decision performance for \citet{yu2018transition} by running the code of the authors for three runs and taking the
average.\footnote{\url{https://github.com/yunan4nlp/NNDisParser}.} We
also present the reported scores for the other top-down RST parsers
\cite{zhang-etal-2020-top,kobayashi2020top}.\footnote{Neither
  \citet{zhang-etal-2020-top} nor \citet{kobayashi2020top} released
  their code, so we were unable to rerun their models.} Human performance in \tabref{mainresult} and \tabref{mainresult2} is the score of human agreement reported by \citet{joty-etal-2015-codra} ad \citet{morey-etal-2017-much}.
%Comparing the reported and reproduced
%performance for the dynamic oracle model, we notice the \texttt{Full}
%performance decreases from 59.9 to 59.3, showing there is some
%variability in the output. All experiments henceforth are run with three
%trials, and the mean performance is presented.

\begin{table}[t]
	\begin{center}
		\begin{adjustbox}{max width=0.95\linewidth}
			\begin{tabular}{lrrrr}
				\toprule
				\textbf{Method} & \textbf{S} & \textbf{N} & \textbf{R} & \textbf{F} \\
				\midrule
				\multicolumn{5}{l}{\textit{Bottom Up: }}\\
				\citet{feng-hirst-2014-linear}*\sentfeat & 68.6 & 55.9 & 45.8 & 44.6 \\
				\citet{ji-eisenstein-2014-representation}*\sentfeat & 64.1 & 54.2 & 46.8 & 46.3 \\
				\citet{surdeanu-etal-2015-two}*\sentfeat & 65.3 & 54.2 & 45.1 & 44.2 \\
				\citet{joty-etal-2015-codra}* &  65.1 & 55.5 & 45.1 & 44.3\\
				\citet{hayashi-etal-2016-empirical}* & 65.1 & 54.6 & 44.7 & 44.1\\
				\citet{li-etal-2016-discourse}* & 64.5 & 54.0 & 38.1 & 36.6\\
				\citet{braud-etal-2017-cross-lingual}* & 62.7 & 54.5 & 45.5 & 45.1\\
				\citet{yu2018transition} (static)\parafeat & 71.1 & 59.7 & 48.4 & 47.4 \\
				\citet{yu2018transition} (dynamic)\parafeat & 71.4 & 60.3 & 49.2 & 48.1 \\
				\midrule
				\multicolumn{5}{l}{\textit{Top Down: }}\\
				\citet{zhang-etal-2020-top}* & 67.2 & 55.5 & 45.3 & 44.3 \\
				%\citet{kobayashi2020top}*$\dagger\ddagger$ & --- & --- & --- & 48.5 \\
				\midrule
				\multicolumn{5}{l}{\textit{Our model}}\\
				Transformer (static)\parafeat & 70.6 & 59.9 & 50.6 & 49.0\\
				Transformer (dynamic)\parafeat & 70.2 & 60.1 & 50.6 & 49.2\\
				LSTM (static)\parafeat & 72.7 & 61.7 & 50.5 & 49.4 \\
				LSTM (dynamic)\parafeat & \textbf{73.1} & \textbf{62.3} & \textbf{51.5} & \textbf{50.3} \\
				\midrule
				\multicolumn{5}{l}{\textit{Our best model without paragraph boundary feature}}\\
				LSTM (static) & 66.3 & 56.6 & 47.1 & 46.1 \\
				LSTM (dynamic) & 67.3 & 57.4 & 48.5 & 47.4 \\
				\midrule
				Human & 78.7 & 66.8 & 57.1 & 55.0 \\
				\bottomrule
			\end{tabular}
		\end{adjustbox}
	\end{center}
	\caption{\label{tab:mainresult} Results over the test set calculated using  micro-averaged F-1 on labelled attachment decisions (original Parseval).
		All metrics (S: \texttt{Span}, N: \texttt{Nuclearity}, R: \texttt{Relation}, F: \texttt{Full}) are averaged over three runs. ``*'' denotes reported performance. ``\sentfeat'' and ``\parafeat'' denote that the model uses sentence and paragraph boundary features, respectively. In this evaluation, \citet{kobayashi2020top} does not report the original Parseval result. }
\end{table}

Overall, in \tabref{mainresult} our top-down models (LSTM and transformer) outperform all bottom-up and top-down baselines across all metrics.  As we saw in the feature addition study,
the LSTM model outperforms the transformer model, even though the
transformer uses pre-trained BERT. We hypothesize that this may be
because BERT is trained over shorter texts (paragraphs or sentence pairs),
while our documents are considerably longer. Also, due to memory
constraints, we break long documents into partial documents
(\secref{transformer}), limiting fine-grained word-to-word attention to
only nearby EDUs.

In \tabref{mainresult}, we also present results for our model without
paragraph features, and compare against other models which don't use
paragraph features (each marked with
``\parafeat'').\footnote{\citet{yu2018transition} use paragraph boundary
  features in their original code, but do not report it in the paper.}
First, we observe that our best model substantially outperforms all
models with paragraph boundary features in terms of the
\texttt{Full} metric.
% including \citet{kobayashi2020top} by +1.8
Compared to \citet{zhang-etal-2020-top}, our models
(without this feature) achieve an improvement of $+0.1$, $+1.9$,
$+3.2$, and $+3.1$ for \texttt{Span}, \texttt{Nuclearity},
\texttt{Relation}, and \texttt{Full} respectively.

%Additionally, we report RST-Parseval result in \tabref{mainresult2}.\footnote{Performance looks unreasonably high with RST-Parseval, and \citet{morey-etal-2017-much} encourage to use original Parseval as the standard evaluation.} 

%As we know, around 77\% paragraphs in RST-DT are actually text spans, and these boundaries are closely related to gold span boundaries in evaluation, which can explicitly lower the difficulty of the task itself

%With regards to the \texttt{Full}
%metric, our proposed LSTM model sets a new state-of-the-art for RST parsing.

\section{Analysis}
\label{sec:analyis}

In \tabref{analisis1} we present the impact of the dynamic oracle over documents of differing length for the LSTM model.  Generally, we found that the static model performs better for shorter documents, and the dynamic oracle is more
effective for longer documents.  For instance, for documents with
50--100 EDUs, the dynamic oracle
improves the \texttt{Span}, \texttt{Nuclearity}, and
\texttt{Relation} metrics substantially. We also observe that the longer the document, the more difficult the tree prediction is. It is confirmed by the decreasing trends of all metrics for longer documents in \tabref{analisis1}.

\begin{table}[t]
	\begin{center}
		\begin{adjustbox}{max width=1\linewidth}
			\begin{tabular}{rcclrrr}
				\toprule
				\textbf{\#EDUs} & \textbf{\#Docs} & \textbf{\#Spans} & \textbf{Type} & \textbf{S} & \textbf{N} & \textbf{R} \\%& \textbf{F} \\
				\midrule
				\multirow{2}{*}{$(0,50]$} & \multirow{2}{*}{21} &\multirow{2}{*}{404} & Static & 81.0 & 72.0 & 58.9 \\%& 58.2 \\
				& & & Dynamic & 79.3 & 71.6 & \textbf{59.1} \\%& 58.3\\
				\midrule
				\multirow{2}{*}{$(50,100]$} & \multirow{2}{*}{9} &\multirow{2}{*}{639} & Static & 76.8 &  66.4 & 56.2 \\%& 54.7 \\
				& & & Dynamic & \textbf{79.3} & \textbf{69.2} & \textbf{59.2} \\%& 57.4\\
				\midrule
				\multirow{2}{*}{$(100,150]$} & \multirow{2}{*}{5} &\multirow{2}{*}{604} & Static & 69.6 & 58.4 & 49.1 \\%& 47.8 \\
				& & & Dynamic & \textbf{70.3} & 57.4 & 49.1 \\%& \textbf{47.5}\\
				\midrule
				\multirow{2}{*}{$(150,\infty)$} & \multirow{2}{*}{3} &\multirow{2}{*}{661} & Static & 66.6 & 53.9 & 41.0 \\%& 40.3 \\
				& & & Dynamic & 66.0 & \textbf{54.4} & \textbf{41.8} \\%& 41.2\\
				\bottomrule
			\end{tabular}
		\end{adjustbox}
	\end{center}
	\caption{\label{tab:analisis1} Impact of the dynamic oracle over
		documents of differing length.  Scores (micro-averaged F-1 on labelled attachment decisions) are averaged over three
		runs on the test set.}% using LSTM.}
\end{table}

\begin{figure}[]
	\centering
	\includegraphics[width=2in]{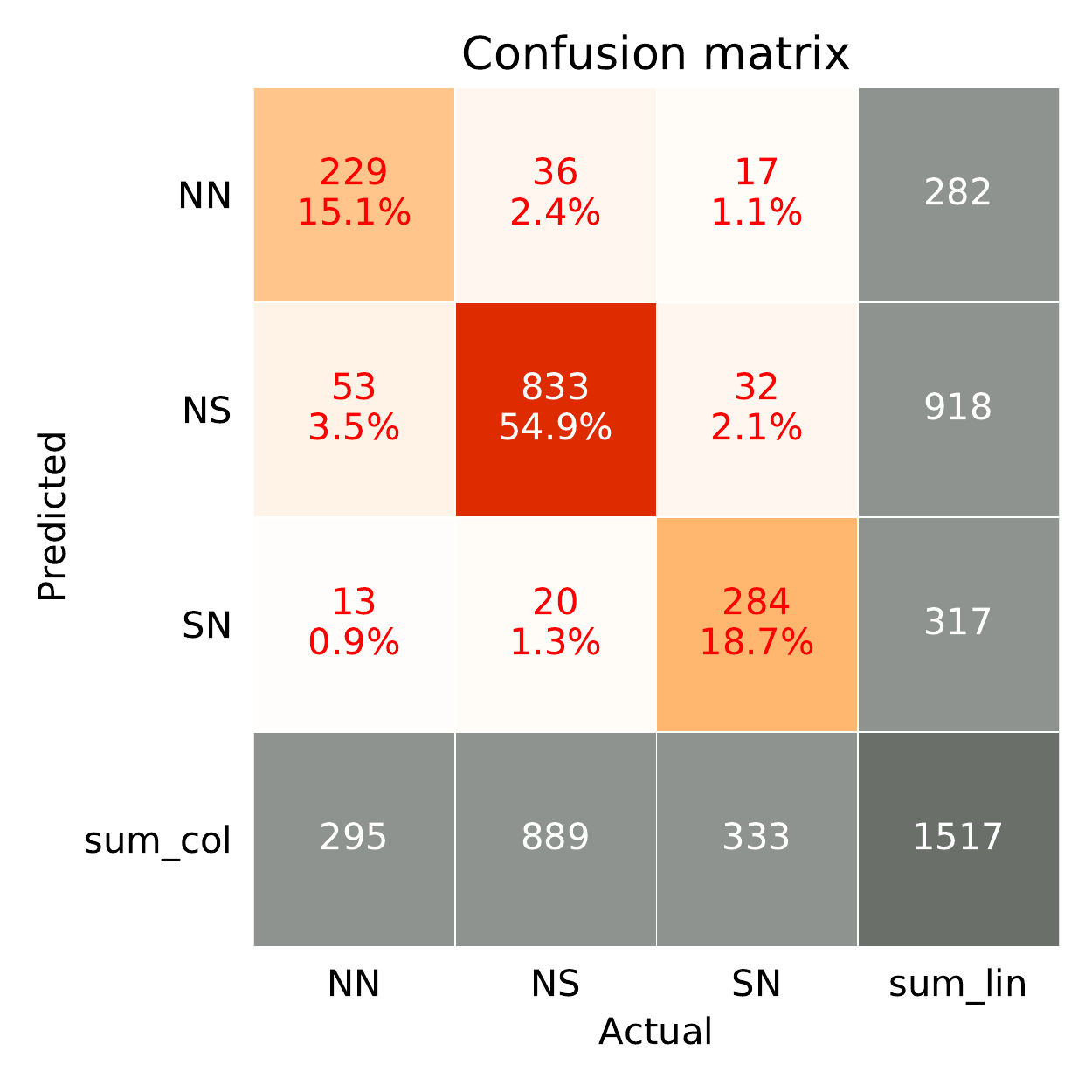}
	\caption{Confusion matrix of nuclearity prediction over the test set (\rstrel{NS} = Nucleus-Satellite, \rstrel{NN} = Nucleus-Nucleus, \rstrel{SN} = Satellite-Nucleus).}
	\label{fig:confusion_nuclear}
\end{figure}

\begin{figure}[t]
	\centering
	\includegraphics[width=1\linewidth]{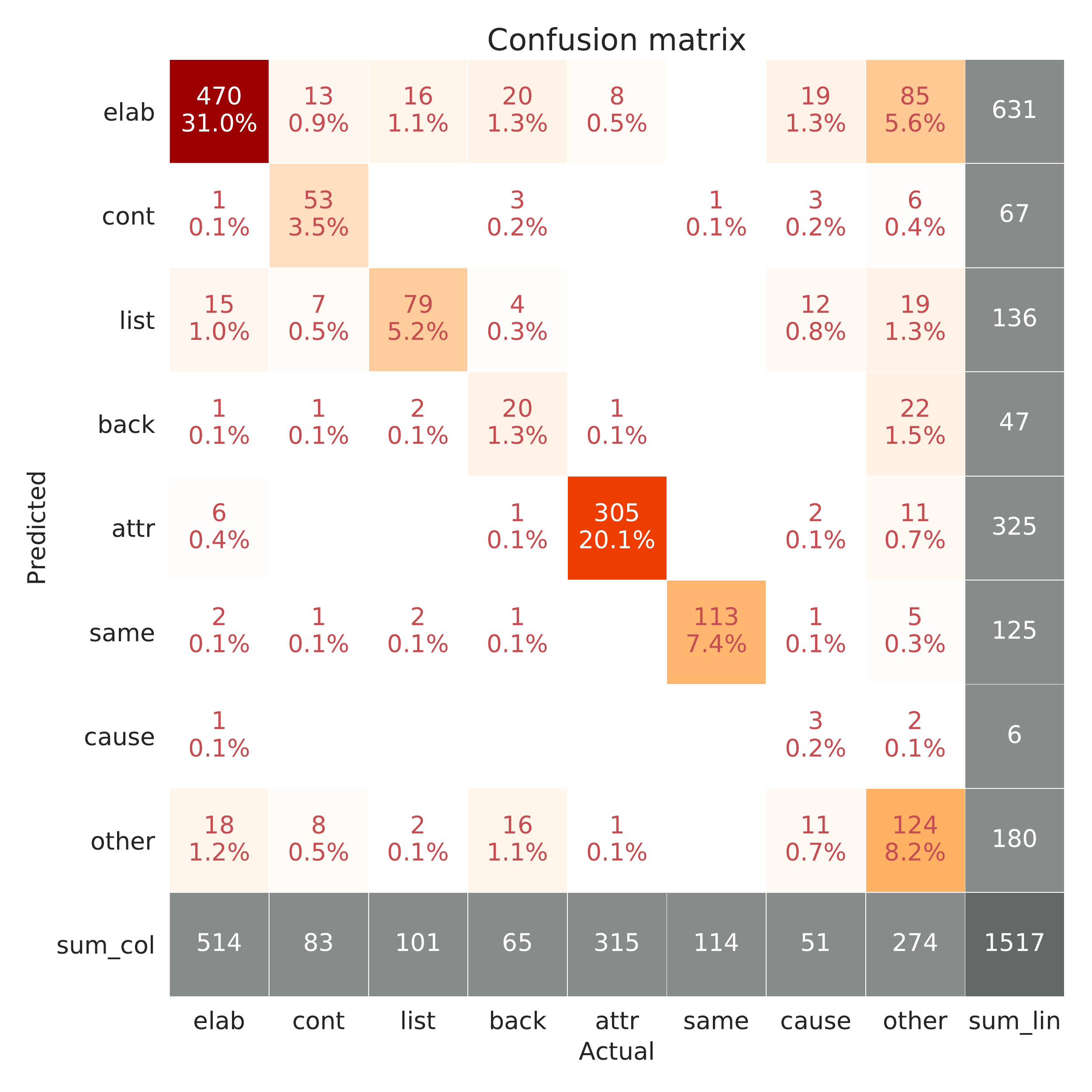}
	\caption{Confusion matrix of relation prediction over the test set with top-7 relations (\rstrel{elab} = Elaboration, \rstrel{cont} = Contrast, \rstrel{list} = List, \rstrel{back} = Background, \rstrel{same} = Same, \rstrel{temp} = Temporal, \rstrel{eval} = Evaluation, \rstrel{other} = Other 11 relations).}
	\label{fig:confusion_relation}
\end{figure}

In total, our best model obtains 1,698 out of 2,308 spans of original Parseval trees, and correctly predict 1,517 segmentation points (pairs). We further analyze these pairs by presenting the confusion matrices of 
nuclearity and relation prediction in \figref{confusion_nuclear} and \figref{confusion_relation}, respectively.

First, the model tends to have more errors on \rstrel{NN} 
(Nucleus--Nucleus) prediction where 53 span pairs (18\% of \rstrel{NN}) 
are classified as \rstrel{NS} (Nucleus--Satellite). Class imbalance in the training set (\rstrel{NN}:\rstrel{NS}:\rstrel{SN} = 23:61:16) is the main factor that drives the model to favor \rstrel{NS} over the other classes.

In \figref{confusion_relation} we present analysis over top-7 relations 
and a relation \rstrel{other} that represents the rest of 
11 classes.
Similar to the nuclearity prediction, the relation class distribution is 
also imbalance where \rstrel{elab} accounts for 37\% of the examples.  
Some relations are related to \rstrel{elab} (see \tabref{relation_ex} 
for examples), such as \rstrel{back}, \rstrel{cause}, and \rstrel{list} 
which we see some false positives.  This produces the low precision of 
\rstrel{elab} (74\%).  Unlike \rstrel{elab}, relation \rstrel{attr} is 
also a major class (represents 14\% of the training data) but its precision and 
recall is substantially higher, at 94\% and 96\% respectively, 
suggesting it is less ambiguous.  For \rstrel{other}, its recall is 
45\%, and most of the errors are classified as \rstrel{elab} (31\%).

%Lastly, for the spans that are correctly predicted by the LSTM Dynamic 
%model, we present a
%breakdown of the misclassifications among the top-10 relations in 
%\tabref{analisis2}.
%Overall we find relations such as \texttt{back}, \texttt{evid},
%\texttt{cause}, \texttt{eval}, and \texttt{temp} that require logical 
%reasoning tend to
%have more errors, with a misclassification rate of more than 70\%.

%\begin{table}[ht]
%	\begin{center}
%		\begin{adjustbox}{max width=1\linewidth}
%			\begin{tabular}{ll@{\;\;\;\;\;\;\;\;\;\;\;\;\;\;}ll}
%				\toprule
%				\multicolumn{2}{c}{\textbf{\#Relation}} &
%				\multicolumn{2}{c}{\textbf{Misclassified (\%)}} \\
%				\midrule
%				\texttt{elab}: 575 & \texttt{attr}: 330 & \texttt{elab}: 11.3 & \texttt{attr}: 7.3 \\
%				\texttt{same}: 121 & \texttt{list}: 113 & \texttt{same}: 4.1 & \texttt{list}: 29.2 \\
%				\texttt{cont}: 95 & \texttt{back}: 74 & \texttt{cont}: 44.2 & \texttt{back}: 72.9 \\
%				\texttt{evid}: 62 & \texttt{cause}: 57 &  \texttt{evid}: 79.0 & \texttt{cause}: 94.7  \\
%				\texttt{eval}: 50 & \texttt{temp}: 46 &  \texttt{eval}: 100.0 & \texttt{temp}: 84.8 \\
%				\bottomrule
%			\end{tabular}
%		\end{adjustbox}
%	\end{center}
%	\caption{\label{tab:analisis2} F-1 error rate for the
%		top-10 relation labels for LSTM Dynamic (based on labelled attachment decisions).}
%	% in the test set using our best dynamic LSTM model.}
%	%In total we correctly predict 4005 out of 4616 EDU spans.}
%\end{table}

%\begin{figure}[t]
%	\centering
%	\includegraphics[width=3in]{image/relation_example.pdf}
%	\caption{Examples of misclassified relations.}
%	\label{fig:relation_ex}
%\end{figure}
\begin{table}[t]
	\begin{center}
		\begin{adjustbox}{max width=1\linewidth}
			\begin{tabular}{p{3.4cm}p{3.4cm}cc}
				\toprule
				\textbf{EDU-1} & \textbf{EDU-2} & \textbf{Actual} &\textbf{Pred} \\
				\midrule
				senator sasser of tennessee is chairman of the appropriations subcommittee on military construction; & mr. bush 's \$ 87 million request for tennessee increased to \$ 109 million.&\rstrel{back}&\rstrel{elab}\\
				\midrule
				a law went in the books in january that let him smoke bacon without breeding pigs. & he chased in & \rstrel{cause} & \rstrel{elab} \\
				\midrule
				that 's the rule. & that 's the market. &  \rstrel{list} & \rstrel{elab} \\
				\bottomrule
			\end{tabular}
		\end{adjustbox}
	\end{center}
	\caption{Examples of misclassified relations.}
	\label{tab:relation_ex}
\end{table}

\section{Conclusion}

We introduce a top-down approach for RST parsing via sequence labelling.
Our model is conceptually simpler than previous top-down discourse parsers and can leverage pre-trained language models such as BERT. We additionally propose a dynamic-oracle for our top-down parser, and demonstrate that our best model achieves a new
state-of-the-art for RST parsing.

%We found that our LSTM model works better than the transformer model, but contend that efforts to leverage these large pre-trained models for the task are preliminary.

%Overall the proposed LSTM works better than the transition based
%approach for all metrics. We also show that the notion
%1) to penalize the model based on the number of elements and;
%2) to apply dynamic oracle can improve the top-down RST parsing.
%However, using pre-trained encoders such as BERT remains challenging in
%terms of memory and generalization. In the future, we aim to find a
%better strategy to fit pre-trained-encoder for the task such as
%discourse parsing.

\section*{Acknowledgments}

We are grateful to the anonymous reviewers for
their helpful feedback and suggestions. In this research, the first author is
supported by the Australia Awards Scholarship (AAS), funded by the Department of Foreign Affairs and Trade (DFAT), Australia.
This research was undertaken using the LIEF HPC-GPGPU Facility hosted at The University of Melbourne. This facility was established with the assistance of LIEF Grant LE170100200.

\bibliography{anthology,acl2020}
\bibliographystyle{acl_natbib}

\clearpage

\appendix

\section{Evaluation with RST-Parseval Procedure}
\label{sec:appendix_res}
\begin{table}[h]
	\begin{center}
		\begin{adjustbox}{max width=0.9\linewidth}
			\begin{tabular}{lrrrr}
				\toprule
				\textbf{Method} & \textbf{S} & \textbf{N} & \textbf{R} & \textbf{F} \\
				\midrule
				\multicolumn{5}{l}{\textit{Bottom-Up}}\\
				\citet{feng-hirst-2014-linear}*\sentfeat & 84.3 & 69.4 & 56.9 &
				56.2 \\
				\citet{ji-eisenstein-2014-representation}*\sentfeat & 82.0 & 68.2 &
				57.8 & 57.6 \\
				\citet{surdeanu-etal-2015-two}*\sentfeat & 82.6 & 67.1 & 55.4 &
				54.9 \\
				\citet{joty-etal-2015-codra}* & 82.6 & 68.3 & 55.8 &
				54.4 \\
				\citet{hayashi-etal-2016-empirical}* & 82.6 & 66.6 & 54.6 & 54.3 \\
				\citet{li-etal-2016-discourse}* & 82.2 & 66.5 & 51.4 & 50.6 \\
				\citet{braud-etal-2017-cross-lingual}* & 81.3 & 68.1 & 56.3 & 56.0 \\
				\citet{yu2018transition} (1 run)*\parafeat & 85.5 & 73.1 & 60.2 & 59.9 \\
				\citet{yu2018transition} (static)\parafeat  & 85.8 & 72.6 & 59.5 & 59.0 \\
				\citet{yu2018transition} (dynamic)\parafeat & 85.6 & 72.9 & 59.8 & 59.3 \\
				\midrule
				\multicolumn{5}{l}{\textit{Top-Down}}\\
				%\citet{zhang-etal-2020-top} & - & - & - & - \\
				\citet{kobayashi2020top}*\sentfeat\parafeat & \textbf{87.0} & \textbf{74.6} & 60.0 & - \\
				\midrule
				\multicolumn{5}{l}{\textit{Our model}}\\
				Transformer (static)\parafeat & 85.2 & 72.0 & 60.3 & 59.6 \\
				Transformer (dynamic)\parafeat & 85.5 & 72.3 & 60.5 & 59.9 \\
				LSTM (static)\parafeat & 86.4 & 73.4 & 60.8 & 60.3 \\
				LSTM (dynamic)\parafeat & 86.6 & 73.7 & \textbf{61.5} & \textbf{60.9} \\
				\midrule
				\multicolumn{5}{l}{\textit{Our best model without boundary feature}}\\
				LSTM (static) & 83.2 & 70.4 & 58.4 & 57.9 \\
				LSTM (dynamic) & 83.6 & 70.4 & 58.8 & 58.2 \\
				\midrule
				Human & 88.3 & 77.3 & 65.4 & 64.7 \\
				\bottomrule
			\end{tabular}
		\end{adjustbox}
	\end{center}
	\caption{\label{tab:mainresult2} 
		Results over the test set calculated using  micro-averaged F-1 on RST-Parseval.
		All metrics (S: \texttt{Span}, N: \texttt{Nuclearity}, R: \texttt{Relation}, F: \texttt{Full}) are averaged over three runs. ``*'' denotes reported performance. ``\sentfeat'' and ``\parafeat'' denote that the model uses sentence and paragraph boundary features, respectively. In this evaluation, \citet{zhang-etal-2020-top} does not report the RST-Parseval result. 
		Also, both \citet{zhang-etal-2020-top,kobayashi2020top} do not release their code. 
		First, We can see that our model achieves the best results in terms of \texttt{Relation} and \texttt{Full}. Compared to other
		models without paragraph boundary features, our proposed model also
		performs best on the \texttt{Full} metric by a comfortable margin. %\citet{kobayashi2020top} results is the highest for \texttt{Span} in this Table but it is likely due to the direct utility of gold sentence and paragraph boundaries for discourse tree construction.
	}
\end{table}

%\vfill\eject
\section{Feature Addition on the Test Set}
\label{sec:appendix_res2}
\begin{table}[h!]
	\begin{center}
		\begin{adjustbox}{max width=0.8\linewidth}
			\begin{tabular}{lcc}
				\toprule
				\textbf{Variants} &  \textbf{LSTM} &
				\textbf{Transformer} \\
				\midrule
				Vanilla &    46.5$\pm$0.4 & 48.0$\pm$0.4 \\
				$+$Syntax &  48.2$\pm$0.7 & 49.3$\pm$1.0  \\
				$+$Penalty & 46.8$\pm$0.8 & \textbf{49.0$\pm$0.2}\\
				$+$Syntax$+$Penalty &  \textbf{49.4$\pm$0.4} & 48.7$\pm$0.6 \\
				\bottomrule
				
			\end{tabular}
		\end{adjustbox}
	\end{center}
	\caption{Feature addition over the test set to find the best configuration for our models. Presented results are the mean and standard deviation of the \texttt{Full} metric (micro-averaged F-score on labelled attachment decisions) over three runs. }
	\label{tab:analysis_test}
\end{table}

\vfill\eject
\section{Model Configuration for Training}
\label{sec:appendix}

\begin{table}[h]
  \centering
  \begin{adjustbox}{max width=1\linewidth}
    \begin{tabular}{lc}
      \toprule
      \textbf{Configuration} & \textbf{Value} \\
      \midrule
      LSTM1, LSTM2, LSTM3 & 200, \textbf{256} \\
      LSTM4 & 100, \textbf{128}, 200 \\
      Word embedding & \textbf{200} \\
      POS embedding & \textbf{200} \\
      EDU type embedding & \textbf{100} \\
      Syntax Feature & \textbf{1200} \\
      $\lambda_1$ & \textbf{1.0}, 1.2 \\
      $\lambda_2$ & 0.6, 0.8, \textbf{1.0}, 1.2 \\
      $\beta$ (Loss penalty) & 0, 0.2, \textbf{0.35}, 0.4, 0.6, 0.8, 1.0 \\
      $\alpha$ (Dynamic oracle) & 0, 0.25, 0.5, \textbf{0.65}, 0.75, 1.0 \\
      Batch size & 2, \textbf{4} \\
      Gradient accumulation & \textbf{2}, 4 \\
      Learning rate & \textbf{0.001} \\
      Dropout probability & \textbf{0.5} \\
      Infrastructure & 1 GPU V100 (16 GB) \\
      Metrics to evaluate & \texttt{Full} \\
      \bottomrule
    \end{tabular}
  \end{adjustbox}
  \caption{\label{app1} Parameter trials of our LSTM model. Bold
    indicates the best value after tuning over the development set. }
\end{table}

\begin{table}[h]
  \centering
  \begin{adjustbox}{max width=1\linewidth}
    \begin{tabular}{lc}
      \toprule
      \textbf{Configuration} & \textbf{Value} \\
      \midrule
      BERT encoder & BERT-Base\\
      Transformer2 & L=2, H=8, FF=2048 \\
      EDU type embedding & \textbf{100} \\
      $\lambda_1$ & 0.5, \textbf{1.0}, 1.5 \\
      $\lambda_2$ & 0.6, 0.8, \textbf{1.0}, 1.5 \\
      $\beta$ (Loss penalty) & 0, 0.2, \textbf{0.35}, 0.4, 0.6, 0.8, 1.0 \\
      $\alpha$ (Dynamic oracle) & 0, 0.25, 0.5, \textbf{0.65}, 0.75, 1.0 \\
      Batch size & 1, \textbf{2} \\
      Gradient accumulation & 2, 3, \textbf{4}, 8 \\
      Learning rate & \textbf{5e-5} \\
      Dropout probability & 0.1, 0.3, 0.4, \textbf{0.5} \\
      Infrastructure & 4 GPU V100 (16 GB) \\
      Metrics to evaluate & \texttt{Full} \\
      \bottomrule
    \end{tabular}
  \end{adjustbox}
  \caption{\label{app2} Parameter trials of our transformer model. Bold
    indicates the best value after tuning over the development set. }
\end{table}

\begin{table}[!h]
  \centering
  \begin{adjustbox}{max width=1\linewidth}
    \begin{tabular}{lcc}
      \toprule
      \textbf{Method} & \textbf{1 Epoch} & \textbf{Converge at Epoch}\\
      \midrule
      Transition-based & 5 mins & 60--70 \\
      LSTM (Ours) & 1 min 21 secs & 60--70 \\
      Transformer (Ours) & 1 min & 130-150 \\
      \bottomrule
    \end{tabular}
  \end{adjustbox}
  \caption{\label{app3} Running time of the static models during the training. The transition-based model is \citet{yu2018transition}}
\end{table}

%\section{Appendices}
%\label{sec:appendix}

\end{document}